\documentclass[journal]{IEEEtran}

\ifCLASSINFOpdf
\else
   \usepackage[dvips]{graphicx}
\fi
\usepackage{url}

\usepackage{graphicx}
\usepackage{amsmath,amssymb,stmaryrd}%ARROWS !

\begin{document}

\title{Direct Text to Speech Translation System using Acoustic Units}

\author{Victoria Mingote, Pablo Gimeno, Luis Vicente, Sameer Khurana, Antoine Laurent, Jarod Duret
\thanks{This work was performed using HPC resources from GENCI–IDRIS (Grants 2022-AD011012565 and AD011012527). This project has received funding from the European Union's Horizon 2020 research and innovation program under the Marie Skłodowska-Curie grant agreement No 101007666. This work was also supported by JSALT 2022 at JHU, with gift-funds from Amazon, Microsoft and Google, and by MCIN/AEI/10.13039/501100011033 and European Union ``NextGenerationEU''/PRTR Grants PDC2021-120846-C41 \& PDI2021-126061OB-C44. }
\thanks{ V. Mingote, P. Gimeno and L. Vicente are with ViVoLab - Arag\'{o}n Institute for Engineering Research (I3A) - University of Zaragoza, Spain, S. Khurana is with MIT Computer Science and Artificial Intelligence Laboratory, Cambridge, MA, USA, A. Laurent is with LIUM - Le Mans University, France, and J.Duret is with LIA - Avignon University, France (e-mail: vmingote@unizar.es, pablogj@unizar.es, lvicente@unizar.es, skhurana@mit.edu, antoine.laurent@univ-lemans.fr, jarod.duret@univ-avignon.fr).}}
%\thanks{S. B. Author, Jr., was with Rice University, Houston, TX 77005 USA. He is now with the Department of Physics, Colorado State University, Fort Collins, CO 80523 USA (e-mail: author@lamar.colostate.edu).}}

\markboth{IEEE Signal Processing Letters}
{Shell \MakeLowercase{\textit{et al.}}: Bare Demo of IEEEtran.cls for IEEE Journals}
\maketitle

\begin{abstract}
This paper proposes a direct text to speech translation system using discrete acoustic units. 
This framework employs text in different source languages as input to generate speech in the target language without the need for text transcriptions in this language. 
Motivated by the success of acoustic units in previous works for direct speech to speech translation systems, we use the same pipeline to extract the acoustic units using a speech encoder combined with a clustering algorithm. 
Once units are obtained, an encoder-decoder architecture is trained to predict them. 
Then a vocoder generates speech from units. 
Our approach for direct text to speech translation was tested on the new CVSS corpus with two different text mBART models employed as initialisation. 
The systems presented report competitive performance for most of the language pairs evaluated. 
Besides, results show a remarkable improvement when initialising our proposed architecture with a model pre-trained with more languages.
\end{abstract}


\IEEEpeerreviewmaketitle

\section{Introduction}
\IEEEPARstart{D}{uring} the last years, the huge increase in the available unlabelled data for text and speech in all languages of the world has led to the need to develop powerful new approaches to process this data.
Also, recent advances in self-supervised learning have provided the opportunity to benefit from this data and produce general-purpose representations.
These representations can be employed for different tasks and languages with impressive results, e.g. for speech processing using XLS-R \cite{Babu21} or for text processing with mBART \cite{liu2020multilingual, li2021multilingual} and mT5 \cite{xue2021mt5}.
%
%This rapid development has  been motivated by the power shown by to create models. 
%
%For speech processing, self-supervised multilingual speech representation learning has evolved to develop large pre-training encoders such as XLS-R \cite{Babu21}. 
%
%Within the same period, several multilingual systems for text data have been created such as mBART \cite{liu2020multilingual, li2021multilingual} and mT5 \cite{xue2021mt5}.
%
Moreover, recently many works have focused on the development of multilingual and also multimodal systems, such as mSLAM \cite{bapna2022mslam}
and SAMU-XLSR \cite{samuxlsr2022}.
These systems aim to reduce communication problems between people speaking and writing different languages, especially in the case of under-resourced languages.

Previous works have established state-of-the-art performance on a variety of  text and speech 
downstream tasks including machine translation, specifically for the text to text and speech to text translation tasks.
However, research interest in speech to speech and text to speech translation tasks is still growing, as these tasks remain a major challenge due to the scarcity of labelled data for fine-tuning the systems. 
%technology 
These tasks seek to convert speech and text generated in a source language into speech in another target language. 
In the case of conventional speech to speech translation, systems rely on a cascade approach that translates speech into text using Automatic Speech Recognition (ASR) followed by text to text machine translation, or a speech to text system.
In both cases, after the mentioned steps, a speech synthesis model is applied to generate speech in the target language. 

The conventional systems mentioned above achieve high performance, but these systems are text-centric.
Thus, having speech in one language as input, an intermediate text representation in the target language has to be obtained as a preliminary step to generate speech.
Therefore, the idea of direct speech to speech translation without relying on intermediate text representation has been recently explored in the literature \cite{jia2022translatotron,jia2019direct}.
%
%This approach has shown greats benefits in terms of lower computational costs and inference latency compared to the cascade approach. 
This approach has shown great computational benefits compared to the cascade approach. 
%
%Nevertheless, a performance gap can still be observed due to the challenges of simultaneously learning the alignment between two languages and the acoustic and linguistic characteristics required to correctly map spectrograms from source to target languages.
Nevertheless, a performance gap can still be observed due to the challenges of simultaneously learning the alignment between two languages and the process of correctly mapping spectrograms from source to target languages.
%
%To tackle the existing gap, the research described in \cite{lee2022direct,lee2021textless} has proposed a system that performs direct speech to speech translation by extracting a set of discrete acoustic units from the target speech.
%
%After that, a speech to unit translation model is trained to predict those discrete representations.
To tackle the existing gap, the research described in \cite{lee2022direct,lee2021textless} has proposed a direct speech to speech translation system which is trained to predict a set of discrete acoustic units extracted from the target speech.
In addition to the direct speech to speech system, these works have introduced a text to speech translation part using discrete acoustic units.
However, these works apply text to unit translation to the output of an ASR system. 
Hence, the proposed approach is not considered a direct text to speech system, as it does not take an original text input directly to produce the output speech.
Moreover, the performance of this system could be influenced by the use of the output of the ASR module as input, the quality of which may affect the subsequent steps.
On the other hand, considering the limitations that still exist in direct translation and the relevance of multimodal and multilingual systems, \cite{duquenne2021multimodal,duquenne2022t} have developed a system for speech to speech and text to speech translation.
In this system, a common fixed representation for speech and text is built to carry out zero-shot cross-modal translation.

Unlike previous works, where text to unit translation systems were used only combined with ASR, this paper describes an implementation of a framework for generating speech in a given language from text input in a different language.
The task can then be formally defined as a direct text to speech translation task.
Applying this framework, we use text as source input to obtain discrete acoustic units as intermediate representations to generate speech. 
Thus, this text framework allows us to generate the same discrete units as using speech as input.
The use of this framework could be useful for different real applications.
For instance, text to speech translation could be employed as a data augmentation technique for low resource languages or to create audio versions of written content, such as podcasts or story-telling services from texts.
%
%Furthermore, in this work, we have also analyzed the effect of using two pre-trained models with a different number of languages as encoder-decoder for the finetuning of our direct text to speech system in the new CVSS dataset.
%
Furthermore, in this work, we have also analyzed the effect of using two pre-trained models with a different number of languages as encoder-decoder for the fine-tuning of our direct text to speech system in a new corpus called Common Voice-based Speech-to-Speech (CVSS) translation \cite{cvss-corpus}. 
This new CVSS dataset has recently been released to address the issues of scarcity in end-to-end labelled data for direct speech to speech and text to speech translation.
In addition, the number of languages in similar previous works has also been limited  to mostly high-resource languages with 10 different languages. 
However, with this new dataset, the text to speech translation task has been evaluated on more than 20 input languages.  

This paper is laid out as follows. 
Section \ref{sec:training} provides a review of the existing approaches which inspire this work, and introduces the proposed direct text to speech framework using acoustic units.
The experimental setup is detailed in Section \ref{sec:exp_setup}, focusing on the data and the evaluation protocol.
Results and discussions are given in Section \ref{sec:task}. 
Finally, conclusions and future lines are presented in Section \ref{sec:conclusion}.

\section{Proposed Method}
\label{sec:training}
\subsection{Preliminaries: Direct Speech to Speech Translation}
Nowadays, there is an expanding line of research in direct speech to speech translation in which the development carried out in \cite{lee2022direct,lee2021textless} has had a great impact.
These works have introduced the first systems based on real speech data as target.
Thus, instead of predicting continuous spectrograms as in \cite{jia2022translatotron,jia2019direct}, discrete units learned from self-supervised representations of the target speech are predicted.
The system proposed is an encoder-decoder based on a sequence-to-sequence transformer model for speech-to-unit translation. 

To create the system described in \cite{lee2022direct,lee2021textless}, two different blocks are integrated.
First, a multilingual Hidden unit BERT (mHuBERT) \cite{hsu2021hubert} is employed to extract representations from the target speech that are then discretized using a quantizer model.
mHuBERT was chosen as generator due to its superior performance across different speech tasks compared to other unsupervised models.
By extracting the discrete units with this approach, the encoder-decoder speech to unit translation model can be trained using the units as target sequence.
In a second step, and once this model is trained, the target speech is generated from the discrete units.

\subsection{Direct Text to Speech Translation}
\paragraph*{Overview.} 
In view of the success achieved by the use of acoustic units for direct speech to speech translation systems in the preliminary works, this work presents a framework to apply the same approach for direct text to speech translation.
On the other hand, the need for multilingual and multimodal systems has also motivated several state-of-the-art translation systems where speech and text are permitted as input.
Therefore, we propose a multilingual framework in which text data is employed as the input source to predict discrete acoustic units as target without the need to know the transcription in the target language.
This aspect is especially relevant in low resource languages, where finding text-speech transcription pairs can be difficult.
In addition, the application of the approach presented in this section can be seen as a data augmentation strategy to be used in the case of these languages with scarcity of available resources.

\begin{figure}[th!]
\includegraphics[width=0.92\linewidth]{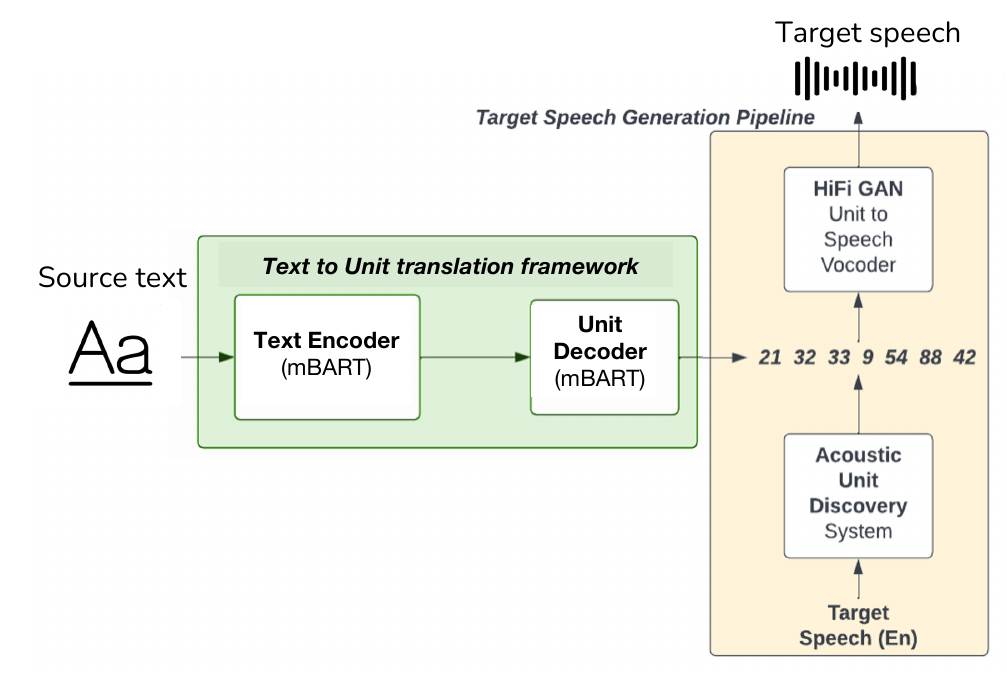}
\caption{Direct text to speech translation system, obtaining acoustic units with source text data in any language to generate target speech in English language.}
\label{fig:textunitsystem}
\end{figure}

As illustrated in Figure~\ref{fig:textunitsystem}, an encoder-decoder architecture is used to perform the direct text to speech translation system. 
Since the conversion of text inputs into acoustic units can be considered as a machine translation task, we have used a pre-trained text model as initialisation for our encoder-decoder architecture. 
Namely, we have considered multilingual BART (mBART) model in its two variations, mBART25 and mBART50 \cite{liu2020multilingual, li2021multilingual}. 
The main difference between both models is the number of languages used in the training process.
After initialisation, the full architecture is fine-tuned on the text to acoustic unit translation task. 
The units employed as targets for this training have previously been extracted with an acoustic unit discovery system.
Finally, in inference, the HiFi GAN \cite{polyak2021speech} unit to speech vocoder is applied to generate target speech utterances.
This unit-based vocoder is a modified version of the original HiFi-GAN neural vocoder presented in \cite{kong2020hifi}. 
For this model, we have used the pre-trained English vocoder available at this link\footnote{\label{note1}\url{https://github.com/facebookresearch/fairseq/blob/main/examples/speech_to_speech/docs/textless_s2st_real_data.md}}.
This last part corresponds to the orange block in Figure~\ref{fig:textunitsystem} and could be shared with a direct speech to speech system.

\paragraph*{Learning.}
To train the direct text to speech translation system, pairs of examples $(x_{S}, u_{L})$ are used where $x_{S}$ is the source text in any of the multiple languages employed, and $u_{L}$ is set of acoustic units extracted from the target speech. 
The generation of these units is carried out by a pre-trained mHuBERT model \cite{lee2021textless} and a k-means quantizer\textsuperscript{\ref{note1}}.
Concerning the mHuBERT model, it is based on the HuBERT Base architecture trained using a combination of English, Spanish, and French data from VoxPopuli \cite{voxpopuli}.
%
%In total, 4.5k hours of data for each language were used.
%
Speech representations are learned in a self-supervised way using unlabelled data as explained in \cite{hsu2021hubert,lakhotia-etal-2021-generative}.
After that, a k-means quantizer is applied to the representations learned in the layer $11th$ of the mHuBERT model to generate discrete labels or units. 
This layer is chosen as done in similar direct translation works \cite{lee2021textless}. 
Several papers have shown that HuBERT like models provide the most meaningful phonetic and word information towards higher layers of the model \cite{hubert_layers1,hubert_layers2}.

To carry out the k-means quantizer process, the two following steps are applied.
First, for training, $N$ centroids are learned using a fraction of the training data.
After that, in inference time, the output of the quantizer is chosen as the index of the centroid minimising the euclidean distance between the input embedding and $N$ centroids learned.
In this case, the number of k-means clusters employed is 1000 as done in \cite{lee2021textless}.
Moreover, the discrete unit sequences extracted from the k-means algorithm could have consecutive repetitions of the same units.
Therefore, to generate the final target units, the original unit sequences are collapsed to convert consecutive equal units into one single unit (e.g., 1 1 2 2 3 3 $\xrightarrow{}$ 1 2 3).
This reduction has been applied since the work described in \cite{lee2022direct} showed that collapsing unit sequences did not lead to a decrease in performance and was more efficient.
As these target units are discrete, the text to unit translation system is trained to minimize the cross entropy loss between the predicted and real units using label smoothing with a probability of $0.2$.
%PABLO: Quiza el tema del label smoothing se puede justificar también con el paper ¨Textless Speech-to-Speech Translation on Real Data" No se si sera ya muy repetitivo 

\paragraph*{Hyperparameters.}
As optimizer for the fine-tuning process, we have employed the Adam optimizer with $\epsilon = 1e-6$, $\beta_{1}=0.9$, $\beta_{2}=0.98$, learning rate $3e-5$, and polynomial learning rate decay scheduling.
The model is trained using the fairseq toolkit \cite{ott2019fairseq} with a dropout of 0.3 and an attention dropout value of 0.1. The training process was carried out employing 8 V100 (32 GB) NVIDIA GPUs.

\section{Experimental Setup}
\label{sec:exp_setup}
\subsection{Data}
\label{sec:data}

For the direct text to speech translation task, two stages have been carried out.
Initially, reference acoustic units are extracted and then text to speech framework is trained using them as targets.
To develop both stages, the following data from the new CVSS translation corpus \cite{cvss-corpus} are employed.
\paragraph*{Acoustic Units.}
For obtaining the acoustic units, the English audios from the CVSS-C (canonical voice) dataset have been used as target speech.
These target audios are forwarded through the acoustic unit discovery system based on mHuBERT model and k-means clustering approach to obtain the discrete unit representations. 
\paragraph*{Direct Text to Speech Translation.}
Once the acoustic units are obtained, they are employed as targets to train the direct text to unit translation system.
Considering that the CVSS dataset also provides the text transcription for the input audios, we have used this dataset to perform 21 languages to EN text to speech translation tasks.

% \subsection{Experimental Description}
\subsection{Evaluation}
\begin{figure}[t]
\includegraphics[width=0.925\linewidth]{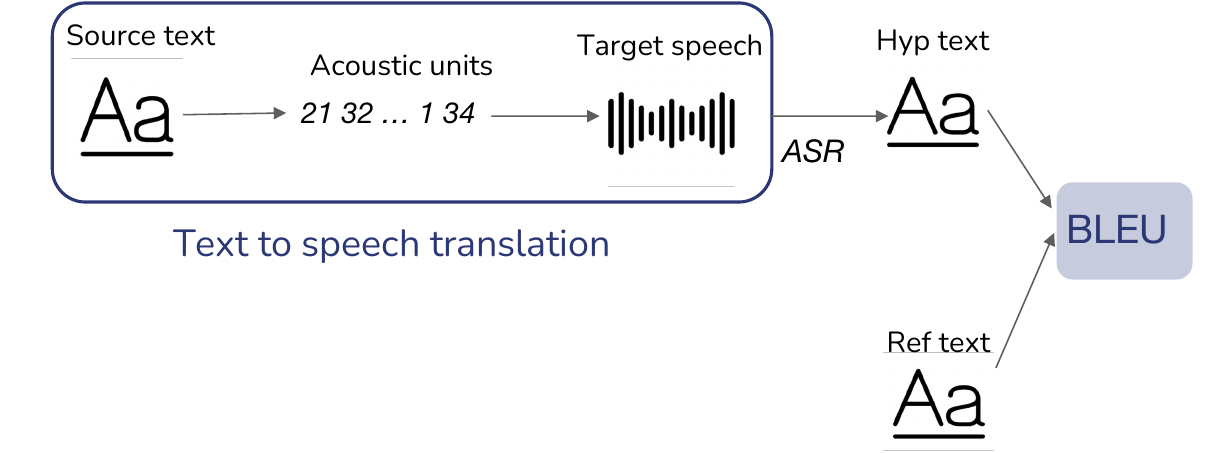}
\caption{Text to speech evaluation pipeline, using an ASR model to generate hypothesis text and compare with reference text to obtain BLEU scores.}
\label{fig:textuniteval}
%\vspace{-0.3cm}
\end{figure}

%Seeking to be able to use the same speech vocoder as the one used in the mentioned speech to speech translation task, we perform the translation in two steps: first a text to acoustic unit conversion and then speech generation through the same HiFi-GAN as described in the previous experiments.
Aiming to evaluate the text to speech translation task, and considering that it is not feasible to directly compare two audio signals, we adopt a similar framework as the one described in \cite{jia2019direct} to evaluate the translation quality of the generated speech. 
This setup is described in Figure \ref{fig:textuniteval}.

As it can be seen, an ASR system is used to generate transcriptions for the target speech. 
The ASR system used\footnote{\url{https://huggingface.co/facebook/wav2vec2-large-960h-lv60-self}} is an open-source English model based on wav2vec 2.0 features trained through a self-training objective \cite{xu2021self}.
The evaluation metric shown in our results is then computed as the BLEU score between the obtained transcriptions and the reference text which is normalized in CVSS to perform this standard evaluation.  
This metric provides an objective measure of speech intelligibility and translation quality.

\section{Results and Discussion}
\label{sec:task}

%\paragraph*{Text to Speech Translation Task.}
As mentioned above, to build the direct text to speech translation system, we have explored different models as initialisation for the encoder-decoder architecture.
Therefore, we have conducted experiments to evaluate the proposed approach using pre-trained mBART25 and mBART50 models.
In addition, we also developed a cascade system in order to have a reference system for comparison. 
This system is composed of a machine translation module based on the mBART50 model followed by a speech synthesis module implemented using tacotron2 \cite{shen2018natural}.

\begin{figure}[h!]  	
    \centering
    \includegraphics[width=0.95\linewidth]{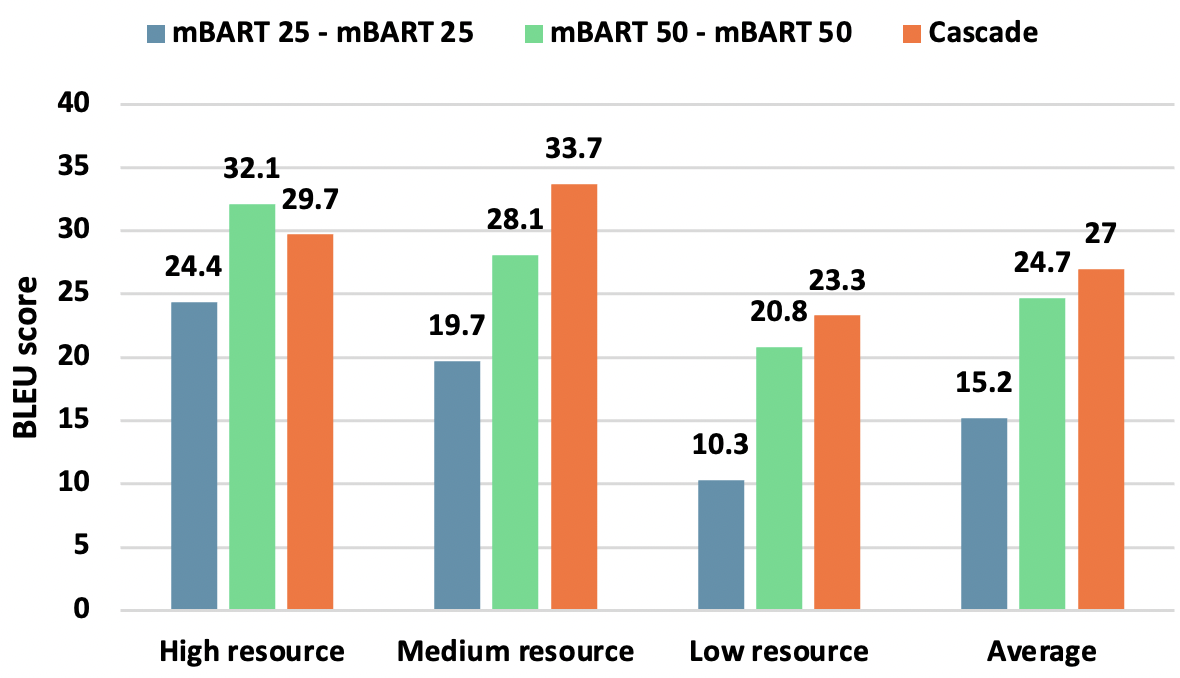}
    \caption{BLEU results on CVSS test set, comparing the cascade and two mBART models used as encoder-decoder initialisation and divided into groups of languages according to the number of resources available for each of them.}
    \label{fig:textunits}
    %\vspace{-0.1cm}
\end{figure}

\begin{figure*}[th!] 
    \centering
    \includegraphics[width=0.95\linewidth]{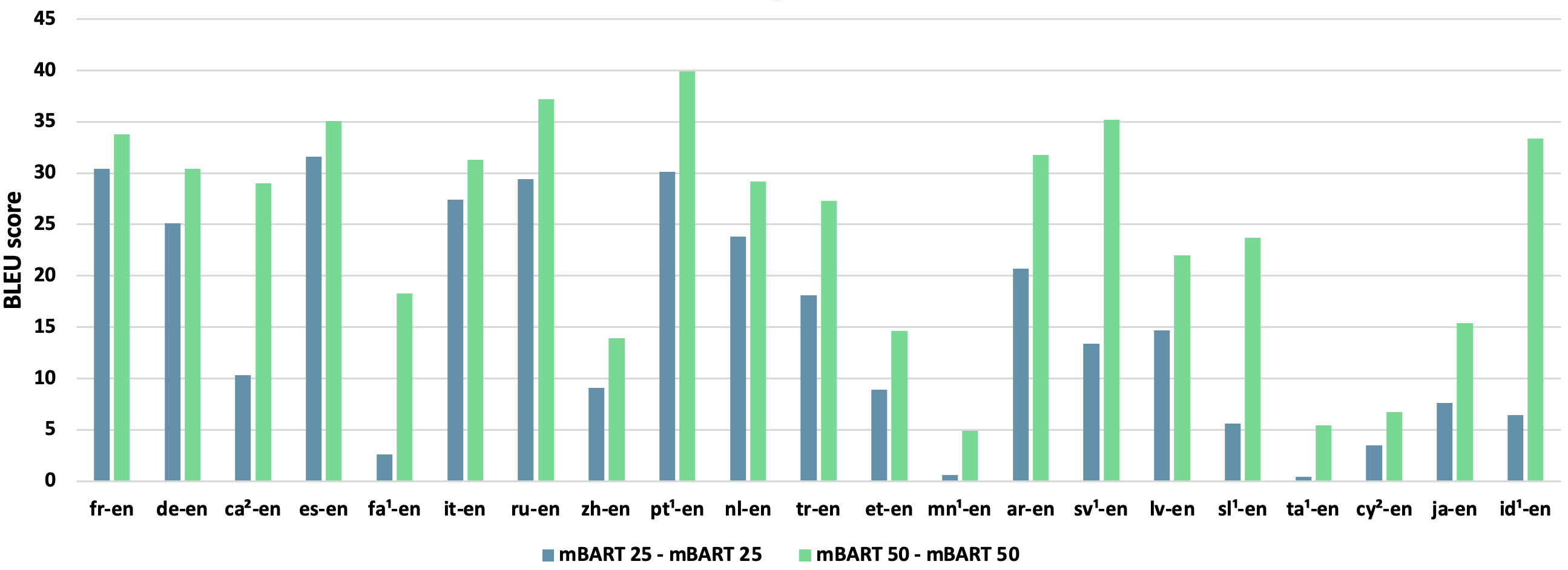}
    %\vspace{-0.1cm}
    \caption{BLEU results on CVSS data test partition for each language available. $^1$ Languages not present in mBART-25, but present in mBART-50. $^2$ Languages not present in mBART-25 or mBART-50.}
    \label{fig:textunitslang}
    %\vspace{-0.4cm}
\end{figure*}

Figure \ref{fig:textunits} presents the BLEU scores in the test partition of the CVSS dataset for our proposed direct text-to-speech system and cascade approach.
In this figure, the performance is shown separately for high, medium and low resource languages.
%In this figure, the performance is shown separately in function of the training data available for high ($> 100h$), medium and low ($< 10h$) resource languages.
%
We have considered high resource languages as those with more than 100h, and low resource languages as those with less than 10h of training data.
Moreover, the average of the results is also presented.
%
%These results show a large performance improvement in all splits when the mBART50 model is used as a pre-training model to initialize our encoder-decoder pipeline for the direct text to speech system.
These results show that the best proposed approach achieves performance close to the cascade system.
Furthermore, our direct text to speech system has the advantage that it does not need to know  the transcription in the target language, while the cascade system needs it to perform the whole translation process.
Note that, if we focus on the two alternatives for the direct text to speech system, a large performance improvement in all splits is observed when the mBART50 model is used as a pre-training model to initialize our encoder-decoder pipeline.

For a more in-depth analysis of the differences found between the two types of mBART models employed, we can see the results for each language of the 21 languages available in the CVSS dataset in Figure \ref{fig:textunitslang}.
This figure shows that the performance of all the languages improves using mBART50.
In addition, the improvement achieved is particularly remarkable in the following translation pairs of languages: fa-en, pt-en, mn-en, sv-en, sl-en, ta-en, id-en, marked with $1$ in the figure.
The relevant performance improvement is motivated by the fact that these languages are not included in mBART25 but are part of the training languages in mBART50.
Note that even languages, such as Catalan (ca) and Welsh (cy) marked with $2$, that are not included in either mBART25 or mBART50, benefit from the influence of having more languages in the second model and improve their results.
To highlight these graphical results, we have calculated the improvement achieved in the three language sets.
We can observe that an average relative improvement of $40\%$ is achieved in terms of BLEU score in the languages employed for the pre-training of both mBART models.
%We can observe that an average improvement of $6.27$ in terms of BLEU points is achieved in the languages employed for the pre-training of both mBART models.
%
In the case of the languages included in mBART50, an average relative improvement of $501\%$ is obtained, while for the languages not present in either of the two, mBART50 achieves an average improvement of $136\%$.
%In the case of the languages included only in mBART50, an average improvement of $14.53$ BLEU points is obtained, while for the languages not present in either of the two, mBART50 achieves an average improvement of $10.95$.
%
These improved results remark the fact that the use of a pre-trained multilingual model in more languages, mBART50, shows a great impact on the results obtained for the new languages included, and also, this increased multilingualism helps to improve the results in languages not presented during the pre-training process.

%\vspace{-0.2cm}
%%%%%%%%%%%%%%%%%%%%%%%%%%%%%%%%% CONCLUSION %%%%%%%%%%%%%%%%%%%%%%%%%%%%%%%
\section{Conclusions and Future Works}
%\vspace{-0.2cm}
\label{sec:conclusion}
In this paper, we have presented a new approach to carry out direct text to speech translation.
This approach is based on an encoder-decoder framework using text as input and discrete acoustic units as the target sequence.
Hence, multilingual text to speech translation can be performed without explicit knowledge of the text transcription in the target language.
The system presented in this paper could be used for different applications such as generating audio books from texts in different languages. 
Moreover, the proposed framework could be applied to get augmented data in order to expand datasets from low resource languages.
%
%As initialisation for the encoder-decoder architecture, we have employed the pre-trained mBART model for language processing.
%
%Moreover, mHuBERT speech encoder combined with k-means has been used to extract these units.
%
The evaluation of this proposal was carried out on the new CVSS dataset to confirm the great performance achieved with this approach to generate speech.
In these experiments, we have also shown an improvement in performance when the model used as initialisation for the encoder-decoder architecture has been pre-trained by including more languages of the translation pairs from the CVSS dataset.
This fact suggests that cross language learning might benefit low resource languages to a significant amount in the text to speech translation task.

The promising results achieved with the proposed system have opened an interesting line of research, so future work will focus on extending our direct text to speech framework to join with a direct speech to speech framework. 
%
%In this way, a multimodal system could build in which speech or text could be used as source input to produce the same discrete acoustic units with either of the 2 modalities and thus generate the target speech.
In this way, a multimodal system could be built in which source input could be speech or text since both modalities are compatible to produce the same discrete acoustic units and thus generate the target speech.
Considering that only speech in the target language is needed, further work could also explore the use of languages different from English as target. %Expandir esto

%\section*{Acknowledgment}

\bibliographystyle{IEEEtran}

\bibliography{mybib}

\begin{thebibliography}{10}
\providecommand{\url}[1]{#1}
\csname url@samestyle\endcsname
\providecommand{\newblock}{\relax}
\providecommand{\bibinfo}[2]{#2}
\providecommand{\BIBentrySTDinterwordspacing}{\spaceskip=0pt\relax}
\providecommand{\BIBentryALTinterwordstretchfactor}{4}
\providecommand{\BIBentryALTinterwordspacing}{\spaceskip=\fontdimen2\font plus
\BIBentryALTinterwordstretchfactor\fontdimen3\font minus
  \fontdimen4\font\relax}
\providecommand{\BIBforeignlanguage}[2]{{%
\expandafter\ifx\csname l@#1\endcsname\relax
\typeout{** WARNING: IEEEtran.bst: No hyphenation pattern has been}%
\typeout{** loaded for the language `#1'. Using the pattern for}%
\typeout{** the default language instead.}%
\else
\language=\csname l@#1\endcsname
\fi
#2}}
\providecommand{\BIBdecl}{\relax}
\BIBdecl

\bibitem{Babu21}
A.~Babu, C.~Wang, A.~Tjandra, K.~Lakhotia, Q.~Xu, N.~Goyal, K.~Singh, P.~von
  Platen, Y.~Saraf, J.~Pino \emph{et~al.}, ``{XLS-R: Self-supervised
  Cross-lingual Speech Representation Learning at Scale},'' in \emph{Proc. ISCA
  Interspeech}, 2022, pp. 2278--2282.

\bibitem{liu2020multilingual}
Y.~Liu, J.~Gu, N.~Goyal, X.~Li, S.~Edunov, M.~Ghazvininejad, M.~Lewis, and
  L.~Zettlemoyer, ``{Multilingual denoising pre-training for neural machine
  translation},'' \emph{Transactions of the Association for Computational
  Linguistics}, vol.~8, pp. 726--742, 2020.

\bibitem{li2021multilingual}
X.~Li, C.~Wang, Y.~Tang, C.~Tran, Y.~Tang, J.~M. Pino, A.~Baevski, A.~Conneau,
  and M.~Auli, ``{Multilingual Speech Translation from Efficient Finetuning of
  Pretrained Models},'' in \emph{Proc. Annual Meeting of the Association for
  Computational Linguistics (Volume 1: Long Papers)}, 2021.

\bibitem{xue2021mt5}
L.~Xue, N.~Constant, A.~Roberts, M.~Kale, R.~Al-Rfou, A.~Siddhant, A.~Barua,
  and C.~Raffel, ``{mT5: A Massively Multilingual Pre-trained Text-to-Text
  Transformer},'' in \emph{Proc. Conference of the North American Chapter of
  the Association for Computational Linguistics: Human Language Technologies},
  2021, pp. 483--498.

\bibitem{bapna2022mslam}
A.~Bapna, C.~Cherry, Y.~Zhang, Y.~Jia, M.~Johnson, Y.~Cheng, S.~Khanuja,
  J.~Riesa, and A.~Conneau, ``{mSLAM: Massively multilingual joint pre-training
  for speech and text},'' \emph{arXiv preprint arXiv:2202.01374}, 2022.

\bibitem{samuxlsr2022}
S.~Khurana, A.~Laurent, and J.~Glass, ``{SAMU-XLSR: Semantically-Aligned
  Multimodal Utterance-level Cross-Lingual Speech Representation},'' \emph{IEEE
  Journal of Selected Topics in Signal Processing}, pp. 1--13, 2022.

\bibitem{jia2022translatotron}
Y.~Jia, M.~T. Ramanovich, T.~Remez, and R.~Pomerantz, ``{Translatotron 2:
  High-quality direct speech-to-speech translation with voice preservation},''
  in \emph{Proc. International Conference on Machine Learning}, 2022, pp.
  10\,120--10\,134.

\bibitem{jia2019direct}
Y.~Jia, R.~J. Weiss, F.~Biadsy, W.~Macherey, M.~Johnson, Z.~Chen, and Y.~Wu,
  ``{Direct Speech-to-Speech Translation with a Sequence-to-Sequence Model},''
  in \emph{Proc. ISCA Interspeech}, 2019, pp. 1123--1127.

\bibitem{lee2022direct}
A.~Lee, P.-J. Chen, C.~Wang, J.~Gu, S.~Popuri, X.~Ma, A.~Polyak, Y.~Adi, Q.~He,
  Y.~Tang \emph{et~al.}, ``{Direct Speech-to-Speech Translation With Discrete
  Units},'' in \emph{Proc. Annual Meeting of the Association for Computational
  Linguistics (Volume 1: Long Papers)}, 2022, pp. 3327--3339.

\bibitem{lee2021textless}
A.~Lee, H.~Gong, P.-A. Duquenne, H.~Schwenk, P.-J. Chen, C.~Wang, S.~Popuri,
  J.~Pino, J.~Gu, and W.-N. Hsu, ``{Textless speech-to-speech translation on
  real data},'' in \emph{Proc. Conference of the North American Chapter of the
  Association for Computational Linguistics: Human Language Technologies},
  2022, pp. 860--872.

\bibitem{duquenne2021multimodal}
P.-A. Duquenne, H.~Gong, and H.~Schwenk, ``{Multimodal and multilingual
  embeddings for large-scale speech mining},'' in \emph{Proc. Advances in
  Neural Information Processing Systems (NeurIPS)}, vol.~34, 2021, pp.
  15\,748--15\,761.

\bibitem{duquenne2022t}
P.-A. Duquenne, H.~Gong, B.~Sagot, and H.~Schwenk, ``{T-Modules: Translation
  Modules for Zero-Shot Cross-Modal Machine Translation},'' in \emph{Proc.
  Conference on Empirical Methods in Natural Language Processing (EMNLP)},
  2022, pp. 5794--5806.

\bibitem{cvss-corpus}
Y.~Jia, M.~T. Ramanovich, Q.~Wang, and H.~Zen, ``{CVSS} corpus and massively
  multilingual speech-to-speech translation,'' in \emph{Proc. Language
  Resources and Evaluation Conference (LREC)}, 2022, pp. 6691--6703.

\bibitem{hsu2021hubert}
W.-N. Hsu, B.~Bolte, Y.-H.~H. Tsai, K.~Lakhotia, R.~Salakhutdinov, and
  A.~Mohamed, ``{Hubert: Self-supervised speech representation learning by
  masked prediction of hidden units},'' \emph{IEEE/ACM Transactions on Audio,
  Speech, and Language Processing}, vol.~29, pp. 3451--3460, 2021.

\bibitem{polyak2021speech}
A.~Polyak, Y.~Adi, J.~Copet, E.~Kharitonov, K.~Lakhotia, W.-N. Hsu, A.~Mohamed,
  and E.~Dupoux, ``{Speech Resynthesis from Discrete Disentangled
  Self-Supervised Representations},'' in \emph{Proc. ISCA Interspeech}, 2021.

\bibitem{kong2020hifi}
J.~Kong, J.~Kim, and J.~Bae, ``{Hifi-gan: Generative adversarial networks for
  efficient and high fidelity speech synthesis},'' in \emph{Proc. Advances in
  Neural Information Processing Systems (NeurIPS)}, vol.~33, 2020, pp.
  17\,022--17\,033.

\bibitem{voxpopuli}
C.~Wang, M.~Riviere, A.~Lee, A.~Wu, C.~Talnikar, D.~Haziza, M.~Williamson,
  J.~Pino, and E.~Dupoux, ``{V}ox{P}opuli: A large-scale multilingual speech
  corpus for representation learning, semi-supervised learning and
  interpretation,'' in \emph{Proc. Annual Meeting of the Association for
  Computational Linguistics (Volume 1: Long Papers)}, 2021, pp. 993--1003.

\bibitem{lakhotia-etal-2021-generative}
K.~Lakhotia, E.~Kharitonov, W.-N. Hsu, Y.~Adi, A.~Polyak, B.~Bolte, T.-A.
  Nguyen, J.~Copet, A.~Baevski, A.~Mohamed, and E.~Dupoux, ``On generative
  spoken language modeling from raw audio,'' \emph{Transactions of the
  Association for Computational Linguistics}, vol.~9, 2021.

\bibitem{hubert_layers1}
A.~Pasad, B.~Shi, and K.~Livescu, ``Comparative layer-wise analysis of
  self-supervised speech models,'' \emph{arXiv preprint arXiv:2211.03929},
  2022.

\bibitem{hubert_layers2}
W.-N. Hsu, Y.-H.~H. Tsai, B.~Bolte, R.~Salakhutdinov, and A.~Mohamed,
  ``{HuBERT: How much can a bad teacher benefit ASR pre-training?}'' in
  \emph{Proc. IEEE International Conference on Acoustics, Speech and Signal
  Processing (ICASSP)}, 2021, pp. 6533--6537.

\bibitem{ott2019fairseq}
M.~Ott, S.~Edunov, A.~Baevski, A.~Fan, S.~Gross, N.~Ng, D.~Grangier, and
  M.~Auli, ``fairseq: A fast, extensible toolkit for sequence modeling,'' in
  \emph{Proc. Conference of the North American Chapter of the Association for
  Computational Linguistics: Human Language Technologies}, 2019, pp. 48--53.

\bibitem{xu2021self}
Q.~Xu, A.~Baevski, T.~Likhomanenko, P.~Tomasello, A.~Conneau, R.~Collobert,
  G.~Synnaeve, and M.~Auli, ``Self-training and pre-training are complementary
  for speech recognition,'' in \emph{Proc. IEEE International Conference on
  Acoustics, Speech and Signal Processing (ICASSP)}, 2021, pp. 3030--3034.

\bibitem{shen2018natural}
J.~Shen, R.~Pang, R.~J. Weiss, M.~Schuster, N.~Jaitly, Z.~Yang, Z.~Chen,
  Y.~Zhang, Y.~Wang, R.~Skerrv-Ryan \emph{et~al.}, ``{Natural tts synthesis by
  conditioning wavenet on mel spectrogram predictions},'' in \emph{Proc. IEEE
  International Conference on Acoustics, Speech and Signal Processing
  (ICASSP)}, 2018, pp. 4779--4783.

\end{thebibliography}

\end{document}